\documentclass[runningheads]{llncs}
\usepackage{graphicx}

\usepackage{tikz}
\usepackage{comment}
\usepackage{amsmath,amssymb} 
\usepackage{color}
\usepackage{bigstrut}
\usepackage{algorithm}
\usepackage{algorithmic}
\usepackage{multirow}
\usepackage{bbm}
\usepackage{booktabs}
\usepackage{subfigure}
\usepackage{enumerate}
\usepackage[misc]{ifsym}
\usepackage{hyperref}

\usepackage[accsupp]{axessibility}  

\begin{document}
\pagestyle{headings}
\mainmatter
\def\ECCVSubNumber{xxxx}  

\title{Acknowledging the Unknown for Multi-label Learning with Single Positive Labels} 


\titlerunning{Acknowledging the Unknown}
%
\author{
Donghao Zhou\inst{1,2} \and
Pengfei Chen\inst{3} \and
Qiong Wang\inst{1} \and 
Guangyong Chen\inst{4}\textsuperscript{(\Letter)} \and
Pheng-Ann Heng\inst{1,5}
}
\authorrunning{Zhou et al.}
%
\institute{
Guangdong Provincial Key Laboratory of Computer Vision and Virtual Reality Technology, Shenzhen Institute of Advanced Technology, \\ Chinese Academy of Sciences, Shenzhen, China \\
\email{dh.zhou@siat.ac.cn} \and
University of Chinese Academy of Sciences, Beijing, China \and
Tencent Technology, Shenzhen, China \and
Zhejiang Lab, Hangzhou, China \\
\email{gychen@zhejianglab.com} \and
The Chinese University of Hong Kong, Hong Kong, China
}

\maketitle

\begin{abstract}
Due to the difficulty of collecting exhaustive multi-label annotations, multi-label datasets often contain partial labels. We consider an extreme of this weakly supervised learning problem, called single positive multi-label learning (SPML), where each multi-label training image has only one positive label. Traditionally, all unannotated labels are assumed as negative labels in SPML, which introduces false negative labels and causes model training to be dominated by assumed negative labels. In this work, we choose to treat all unannotated labels from an alternative perspective, i.e. acknowledging they are unknown. Hence, we propose entropy-maximization (EM) loss to attain a special gradient regime for providing proper supervision signals. Moreover, we propose asymmetric pseudo-labeling (APL), which adopts asymmetric-tolerance strategies and a self-paced procedure, to cooperate with EM loss and then provide more precise supervision. Experiments show that our method significantly improves performance and achieves state-of-the-art results on all four benchmarks. Code is available at \href{https://github.com/Correr-Zhou/SPML-AckTheUnknown}{https://github.com/Correr-Zhou/SPML-AckTheUnknown}.


\end{abstract}

\section{Introduction}
Each image is assumed to associate with only one label in \textit{multi-class} classification, while the real world is closer to being \textit{multi-label}, since multiple objects are inherently contained in a realistic scene \cite{yun2021re}. 
Along with the wide adoption of deep learning, recent years have witnessed great progress in multi-label classification \cite{ridnik2021asymmetric,zhu2021residual,liu2021query2label}. 
Unfortunately, it is excessively laborious to collect exhaustive multi-label annotations for a large-scale image dataset \cite{deng2014scalable}. The major reasons are that the number of potential labels for an image can be large \cite{liu2021emerging} and human annotators tend
to ignore rare or small objects \cite{wolfe2005rare}, which could cause a large amount of label noise. In fact, some publicly available multi-label datasets, including MS-COCO \cite{lin2014microsoft} and Open Images \cite{kuznetsova2020open}, are considered to contain only partial labels of images \cite{huynh2020interactive}. Therefore, it is of scientific interest to explore multi-label learning with partially labeled datasets. 

\begin{figure}[t]
  \centering
    \includegraphics[width=1\linewidth]{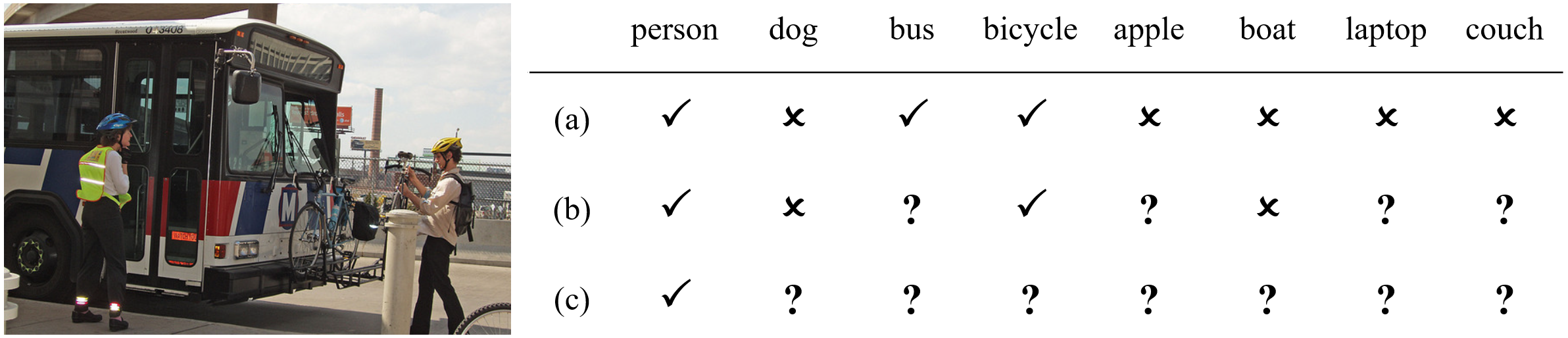}
    \caption{Example of the image with (a) full annotations, (b) partial annotations, and (c) “singe positive” annotations. In single positive multi-label learning (SPML), for this image, only one positive label (\textit{person}) is annotated, and the other labels (\textit{dog}, \textit{bus}, \textit{bicycle}, etc) remain unannotated}
    \label{fig:annotations}
\end{figure}

In this paper, we focus on an extreme of this weakly supervised learning problem in multi-label classification called \textit{single positive multi-label learning} (\textit{SPML}) \cite{cole2021multi}, where only one single positive label (and no other positive or negative labels) is annotated for each training image (see Fig.~\ref{fig:annotations}). 
SPML is a worth-exploring variant of multi-label learning for the following reasons: 
First, this is a common setting in numerous practical scenarios. For instance, when collecting images from the web, the only annotated positive label of an image is from its query \cite{li2017webvision}.
Second, some multi-class datasets like ImageNet \cite{russakovsky2015imagenet} actually contain images that associates with more than one label \cite{beyer2020we,stock2018convnets,tsipras2020imagenet}, while the multi-class training fashion penalizes any predictions beyond the single annotated label \cite{yun2021re}. 
Third, in-depth research of multi-label learning with this minimal supervision could significantly relax the annotation requirement of large-scale multi-label datasets, which helps to reduce expensive annotation costs \cite{durand2019learning}.

Due to the lack of supervision from all negative labels and most positive labels, SPML remains a challenging problem.
Unfortunately, if ignoring unannotated labels, the model trained with only positive labels would collapse to a trivial solution.
Since negative labels are generally the overwhelming majority of multi-label annotations, the traditional method is to assume all unannotated labels are negative to provide supervision signals, which is commonly regarded as the baseline of SPML \cite{cole2021multi}. 
However, this assumption would introduce false negative labels and make model training be dominated
by assumed negative labels, which damages model generalization and causes a substantial performance drop.
Therefore, unannotated labels should be \textit{properly treated} in model training.
In this work, instead of making any unrealistic assumptions, we choose to treat all unannotated labels from a quite different perspective in SPML, i.e. \textit{acknowledging the fact that they are unknown}, and propose a simple but efficient method in response to the issues above.
Our main contributions are summarized as follows:
\begin{enumerate}[1.]
    \item Motivated by the idea of acknowledging the unknown, we propose a novel loss function called \textit{entropy-maximization (EM) loss} for SPML, which aims to maximize the entropy of predicted probabilities for unannotated labels. Besides, we reveal the mechanism of EM loss with gradient-based analysis, which demonstrates that EM loss is capable of tackling inherent issues of the traditional method due to its special gradient regime. 
    \item For more precise supervision, we propose \textit{asymmetric pseudo-labeling (APL)} to cooperate with EM loss. Considering the positive-negative label imbalance of unannotated labels, APL adopts asymmetric-tolerance strategies for positive and negative pseudo-labels with a self-paced procedure, which can generate highly accurate pseudo-labels to further boost performance. 
    \item Experiments performed on four popular multi-label datasets, i.e. PASCAL VOC \cite{everingham2011pascal}, MS-COCO \cite{lin2014microsoft}, NUS-WIDE \cite{chua2009nus}, and CUB \cite{wah2011caltech}, show that our method achieves state-of-the-art results on all four benchmarks. Moreover, detailed ablation study and further analysis verified the effectiveness and rationality of the proposed method.
\end{enumerate}


\section{Related Work}
\noindent \textbf{Multi-label Learning with Weak Supervision.} 
There are several multi-label tasks similar to single positive multi-label learning (SPML). 
Multi-label learning with missing labels (MLML) assumes that some labels are annotated and the others are “missing” for each training image \cite{sun2010multi}.
Different methods have been proposed for MLML, including treating missing labels as negative ones \cite{wang2016cnn,mahajan2018exploring,wang2014binary,sun2017revisiting}, label matrix completion \cite{cabral2011matrix,xu2013speedup,cabral2014matrix}, learning image-label similarities \cite{huynh2020interactive}, etc. 
Semi-supervised multi-label learning (SSML) assumes training data is a subset of images with exhaustive labels and a large number of images without any labels \cite{liu2021emerging}.
Common solutions to SSML include non-negative matrix factorization \cite{liu2006semi}, label propagation \cite{wang2018adaptive}, aligning image features \cite{wang2020dual}, etc.
However, these two tasks are quite different from SPML, where all training images are labeled but each one only contains one positive label. 
Within the limit of this special setting, most standard methods of these two tasks, like consistency regularization \cite{NIPS2016_30ef30b6} and learning label correlations \cite{durand2019learning,zhu2017multi}, are not applicable to SPML.

\noindent \textbf{Entropy Min-/Maximization.} 
The idea of entropy min-/maximization has been widely exploited in various fields.
Grandvalet et~al. proposed entropy minimization regularization for semi-supervised learning \cite{grandvalet2004semi}, and Pereyra et~al. have shown that penalizing low entropy also acts as a strong regularizer in multi-class classification \cite{pereyra2017regularizing}. 
In domain adaptation, entropy min-/maximization has been adopted to adversarially optimize the model \cite{Saito_2019_ICCV} and learn from target domain data \cite{wu2020entropy}.
Besides, entropy maximization has been shown to improve exploration in reinforcement learning \cite{mnih2016asynchronous}.
To the best of our knowledge, our entropy-maximization (EM) loss is the \textit{first} loss function to utilize entropy maximization for unannotated labels and demonstrate its effectiveness.

\noindent \textbf{Pseudo-Labeling.} 
The goal of pseudo-labeling is to assign labels to under-labeled samples with a trained model \cite{lee2013pseudo}. 
There are many methods to implement pseudo-labeling, including exploiting neighborhood graphs \cite{Iscen_2019_CVPR}, performing clustering \cite{Taherkhani_2021_CVPR}, estimating prediction uncertainty \cite{durand2019learning,rizve2021defense}, etc. 
However, existing pseudo-labeling techniques fail to consider the positive-negative label imbalance of unannotated labels in SPML, which would result in a large amount of label noise. 
By adopting asymmetric-tolerance strategies for positive and negative pseudo-labels, our asymmetric pseudo-labeling (APL) can significantly reduce wrong pseudo-labels, allowing for better performance.


\section{Methodology}
In this section, we start by giving a formal definition of single positive multi-label learning (SPML) in Sec.~\ref{sec:3.1}. 
Then, we would introduce our entropy-maximization (EM) loss designed for SPML in Sec.~\ref{sec:3.2}. Finally, we would describe the proposed asymmetric pseudo-labeling (APL) in Sec.~\ref{sec:3.3}. 

\subsection{Problem Definition}
\label{sec:3.1}
Let $\mathcal{D}=\{(\mathbf{x}^{(n)},\mathbf{y}^{(n)})\}_{n=1}^{N}$ denotes a partially labeled multi-label dataset with $N$ images, where each image $\mathbf{x}^{(n)}$ from input space $\mathcal{X}$ is associated with a vector of labels $\mathbf{y}^{(n)}$ from the label space $\mathcal{Y} = \{-1, 0, 1\}^C$ with $C$ classes.
Let $y^{(n)}_c$ be the $c$-th entry of $\mathbf{y}^{(n)}$, where $y^{(n)}_c = 1$ indicates $\mathbf{x}^{(n)}$ is relevant to the $c$-th class (i.e. $\mathbf{x}^{(n)}$ contains a positive label of the $c$-th class) and $y^{(n)}_c = -1$ otherwise. Besides, $y^{(n)}_c = 0$ indicates that the label of the $c$-th class is unannotated for $\mathbf{x}^{(n)}$.
In SPML,  $\mathbf{y}^{(n)}$ satisfies $y^{(n)}_c \in \{0, 1\}$ and $\sum^C_{c=1} \mathbbm{1}_{[y^{(n)}_c = 1]} = 1$, where $\mathbbm{1}_{[\cdot]}$ denotes the indicator function, meaning that only one positive label (and no other positive or negative labels) is annotated for each training image.

The goal of SPML is to learn a mapping function $f:\mathcal{X} \to \mathcal{Y}^\prime$ from $\mathcal{D}$, where $\mathcal{Y}^\prime = \{-1, 1\}^C$ is the ground-truth label space for $\mathcal{X}$.
Defining $f$ as a deep model $f(\cdot; \mathbf{w})$ with the weights $\mathbf{w}$, the standard training approach is to treat each label prediction as an independent binary classification and solve
$
  \hat{\mathbf{w}} = \mathop{\arg\min}_{\mathbf{w}}
  \frac{1}{N} \sum_{n=1}^{N} \mathcal{L}(\mathbf{f}^{(n)}, \mathbf{y}^{(n)}),
$
where $\mathbf{f}^{(n)} = f(\mathbf{x}^{(n)}; \mathbf{w}) \in [0,1]^C$ denotes the predicted probabilities for $\mathbf{x}^{(n)}$ activated by the sigmoid function, and $\mathcal{L}: [0,1]^C \times \mathcal{Y} \to \mathbb{R}$ denotes a loss function that can handle labels $\mathbf{y}^{(n)}$. Let $f^{(n)}_c$ denotes the $c$-th entry of $\mathbf{f}^{(n)}$.

\subsection{Entropy-Maximization Loss}
\label{sec:3.2}

\begin{figure*}[t]
    \centering
    \includegraphics[width=1\linewidth]{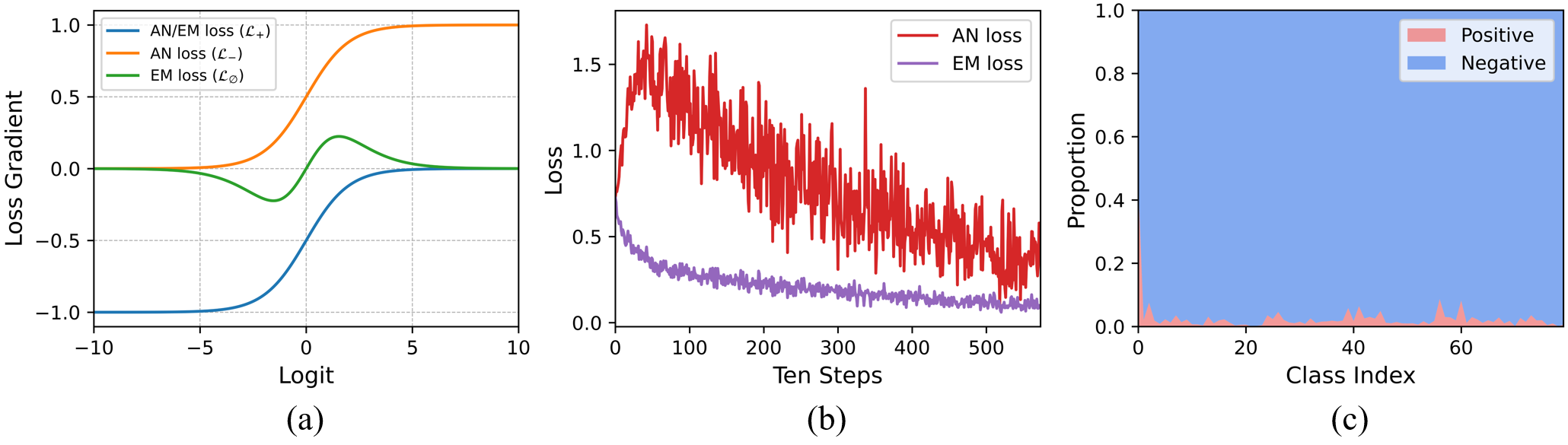}
    \caption{
    (a): Gradients of $\mathcal{L}_{+}$, $\mathcal{L}_{-}$ and $\mathcal{L}_{\varnothing}$. For fair comparison, $\alpha$ of $\mathcal{L}_{\varnothing}$ is set to 1.
    (b): Training losses (averaged in 3 runs) of annotated positive labels (i.e.  $\mathcal{L}_{+}$) on VOC from the model trained with AN and EM loss, where $\alpha$ is also set to 1. Note that $\mathcal{L}_{+}$ of AN loss is more unstable and would increase in early training.
    (c): Proportions of unannotated positive and negative labels of each class on COCO
    }
    \label{fig:method}
\end{figure*}

Since negative labels are generally the overwhelming majority of multi-label annotations, the widely recognized baseline of SPML is \textit{assuming-negative (AN) loss} \cite{cole2021multi}, which assumes all unannotated labels as negative ones for training. Given a image-label pair $(\mathbf{x}^{(n)}, \mathbf{y}^{(n)})$,  AN loss is defined as
\begin{align}
    \mathcal{L}_\text{AN} (\mathbf{f}^{(n)}, \mathbf{y}^{(n)}) = 
    - & \frac{1}{C} \sum_{c=1}^{C}
    [
        \mathbbm{1}_{[y^{(n)}_c = 1]} \log(f^{(n)}_c)
    +   \mathbbm{1}_{[y^{(n)}_c = 0]} \log(1 - f^{(n)}_c)
    ].
    \label{eq:AN}
\end{align}

To better understand the influence of AN loss on model training, we perform gradient-based analysis for AN loss, which is commonly used to deeply study loss functions \cite{lin2017focal,ridnik2021asymmetric}. 
For convenience, 
let $g = g^{(n)}_c$ denotes the output logit of the  $c$-th class for $\mathbf{x}^{(n)}$, 
$p = 1 / (1 + e^{-g})$ (i.e. the sigmoid function) denotes the predicted probability,
and $\mathcal{L}_\mathcal{+} = -\log (p)$ (resp. $\mathcal{L}_\mathcal{-} = -\log (1 - p)$) denotes the loss of the annotated positive label (resp. assumed negative label) in AN loss. 
The gradients of AN loss for the logit $g$ are
\begin{equation}
\left\{
\begin{aligned}
    \frac{\partial \mathcal{L}_{+}}{\partial g} & = 
    \frac{\partial \mathcal{L}_{+}}{\partial p} \frac{\partial p}{\partial g} =
    \frac{-e^{-g}}{1 + e^{-g}}, 
    \quad  y^{(n)}_c = 1, \\
    \frac{\partial \mathcal{L}_{-}}{\partial g} & = 
    \frac{\partial \mathcal{L}_{-}}{\partial p} \frac{\partial p}{\partial g} =
    \frac{1}{1 + e^{-g}}, 
    \quad y^{(n)}_c = 0,
\end{aligned}
\right.
\label{eq:AN_G}
\end{equation}
which shows that AN loss utilizes the same gradient regime for $\mathcal{L}_\mathcal{+}$ and $\mathcal{L}_\mathcal{-}$ (see Fig.~\ref{fig:method}(a)). 
Thus, the model trained with AN loss would treat annotated positive labels and assumed negative labels equally during training, which would result in bad performance in SPML due to the following issues:

\begin{enumerate}[1.]
    \item \textit{Dominance of Assumed Negative Labels}: 
    For each training sample, assumed negative labels are $C-1$ times more than annotated positive ones.
    Hence, affected by the same gradient regime, the model training would be dominated by assumed negative labels until they are well fitted, which hinders the model from learning from annotated positive labels (see Fig.~\ref{fig:method}(b)).
    \item \textit{Introduced Label Noise}: 
    It is unavoidable that there are positive labels wrongly assumed as negative ones in AN loss. Due to the same gradient regime, false negative labels and true positive labels would severely confuse the model.
    \item \textit{Over-Suppression for Confident Positive Predictions}: 
    During training, the model may output a confident positive prediction for an assumed negative label that may be a ground-truth positive one. 
    Unfortunately, AN loss would provide a large gradient for it (see Fig.~\ref{fig:method}(a)), aiming to attain a smaller logit.
\end{enumerate} 

Despite there are some common approaches to reducing the impacts of these issues (e.g. adopting regularization techniques), the gradient regime of AN loss, caused by its unrealistic assumption, would still damage model generalization.
Therefore, unannotated labels need to be properly treated during training, or more specifically, be treated with \textit{a better gradient regime}.
Instead of assuming they are negative, we choose to treat them from an alternative perspective, i.e. acknowledging they are unknown. Motivated by that, we propose \textit{entropy-maximization (EM) loss} for SPML, which is defined as
\begin{align}
    \mathcal{L}_\text{EM} (\mathbf{f}^{(n)}, \mathbf{y}^{(n)}) & = 
    - \frac{1}{C} \sum_{c=1}^{C}
    [
               \mathbbm{1}_{[y^{(n)}_c = 1]} \log(f^{(n)}_c)
    +          \mathbbm{1}_{[y^{(n)}_c = 0]} \alpha H(f^{(n)}_c)
    ],
    \label{eq:EM} \\
    H(f^{(n)}_c) & = -
    [
           f^{(n)}_c      \log(f^{(n)}_c)
        + (1 - f^{(n)}_c) \log(1 - f^{(n)}_c)
    ],
    \label{eq:H}
\end{align}
where $\alpha$ is a hyperparameter to down-weight the strength of entropy maximization.
Note that EM loss adopts entropy maximization as a \textit{separate} loss term (instead of an extra auxiliary regularizer) for unannotated labels to independently provide supervision signals for them,
which is a unique loss design.

Likewise, we reveal the mechanism of EM loss by analyzing its gradients. 
Let $\mathcal{L}_{\varnothing} = \alpha [p\log{p} + (1-p)\log{(1-p)}]$ denotes the loss of the unannotated label in EM loss. 
The gradients of EM loss for the logit $g$ are
\begin{equation}
\left\{
\begin{aligned}
    \frac{\partial \mathcal{L}_{+}}{\partial g} & = 
    \frac{\partial \mathcal{L}_{+}}{\partial p} \frac{\partial p}{\partial g} =
    \frac{-e^{-g}}{1 + e^{-g}}, 
    \quad & y^{(n)}_c = 1, \\
    \frac{\partial \mathcal{L}_{\varnothing}}{\partial g} & = 
    \frac{\partial \mathcal{L}_{\varnothing}}{\partial p} \frac{\partial p}{\partial g} =
    \frac{- \alpha  e^{-g} \log{e^{-g}}}{(1 + e^{-g})^2}, 
    \quad & y^{(n)}_c = 0,
\end{aligned}
\right.
\label{eq:EM_G}
\end{equation}
which shows that EM loss utilizes the same gradient regime as AN loss for annotated positive labels and adopts a quite different one for unannotated labels (see Fig.~\ref{fig:method}(a)). Compared with AN loss, the special gradient regime of EM loss can lead to the following beneficial training behaviours (see Appendix~\ref{sec:B} for additional empirical evidence of these claims):
\begin{enumerate}[1.]
    \item \textit{Learning from Annotated Labels Preferentially}: 
    In early training, the model would produce ambiguous predictions before being well trained, i.e. predicted probabilities (resp. output logits) would be near 0.5 (resp. 0). 
    At this time, the gradients of $\mathcal{L}_{+}$ (resp. $\mathcal{L}_{\varnothing}$) would be relatively large (resp. small) (see Fig.~\ref{fig:method}(a)), leading the model to preferentially learn from annotated positive labels (see Fig.~\ref{fig:method}(b)). 
    Unlike simply down-weighting $\mathcal{L}_{-}$, 
    EM loss tends to keep the predictions of unannotated labels ambiguous, and thus is capable of providing small gradients for them throughout training.
    \item \textit{Mitigating the Effect of Label Noise}: 
    There are no positive labels wrongly regarded as negative ones in EM loss, which prevents the model from producing incorrect negative predictions. 
    Moreover, though unannotated positive labels still exist, the model trained with EM loss would mainly focus on the annotated ones due to the special gradient regime.
    \item \textit{Maintaining Confident Positive Predictions}: 
    After preferentially learning from annotated positive labels, the model is more likely to output confident positive predictions for potential positive labels. 
    When the logit is large enough, the gradient of $\mathcal{L}_{\varnothing}$ would decline and even approach 0 as the logit goes larger (see Fig.~\ref{fig:method}(a)), which helps to maintain these confident positive predictions.
\end{enumerate}


\noindent \subsection{Asymmetric Pseudo-Labeling}
\label{sec:3.3}

\begin{algorithm}[tb]
\caption{Asymmetric Pseudo-Labeling}
\textbf{Input}: Training set $\mathcal{D}$ and model $f_{T_\text{w}}$ trained with Eq.~\ref{eq:EM} for $T_\text{w}$ epochs \\
\textbf{Parameter}: Total training epoch $T_\text{t}$, sample proportion $\theta\%$ and loss weight $\beta$  \\
\textbf{Output}: Well-trained model $f_i$
\begin{algorithmic}[1] 
\STATE $i \leftarrow T_\text{w}$, $\theta^\prime\% \leftarrow \theta\% / (T_\text{t} - T_\text{w})$
\REPEAT 
\STATE Generate pseudo-labels using $f_{i}$ by following Eq.~\ref{eq:PL}
\STATE Train $f_{i+1}$ from $f_{i}$ with Eq.~\ref{eq:EM_PL}
\STATE $i \leftarrow i+1$
\UNTIL early stopping \OR $i = T_\text{t}$
\STATE \textbf{return} $f_i$
\end{algorithmic}
\label{alg:PL}
\end{algorithm}

Exploiting a better gradient regime, the model trained with EM loss can produce more distinguishable predictions for unannotated positive and negative labels (see Fig.~\ref{fig:anal}(a)\&(b)). 
According to these informative predictions, we can assign pseudo-labels to partially labeled images, aiming to provide more precise supervision for the model and then further improve its performance. 
Generally, pseudo-labeling is performed by setting a predefined rule to select potential positive or negative labels, e.g. setting a score threshold \cite{lee2013pseudo} or sample proportion \cite{chen2020self}. 
In pseudo-labeling, there is a natural trade-off between the provided supervision and the introduced noise, since generating more pseudo-labels is often accompanied by introducing more noise.
Specifically, adopting a \textit{low-tolerance} strategy (high score threshold or low sample proportion) would provide less accurate pseudo-labels, whereas adopting a \textit{high-tolerance} strategy (low score threshold or high sample proportion) would introduce more noisy pseudo-labels. 

Different from the other tasks, in SPML, the amounts of unannotated positive and negative labels are quite imbalanced on most classes (see Fig.~\ref{fig:method}(c), refer to Appendix~\ref{sec:F.1} for more examples).
Considering this enormous imbalance, we choose to treat positive and negative pseudo-labels with \textit{asymmetric tolerance}
and thus propose \textit{asymmetric pseudo-labeling (APL)} to cooperate with EM loss, 
aiming to generate relatively sufficient and accurate pseudo-labels.
For negative pseudo-labels, APL adopts a \textit{high-tolerance} strategy with a high sample proportion $\theta\%=90\%$, which means that $90\%$ of unannotated labels on a class could be selected as negative pseudo-labels at most during training.
For positive ones, APL directly ignores them to avoid introducing any noisy positive pseudo-labels that would significantly reduce performance (see Fig.~\ref{fig:ABL}(b)), which can be regarded as an extreme \textit{low-tolerance} strategy.

As shown in Algorithm~\ref{alg:PL}, after warm-up for $T_\text{w}$ epochs, pseudo-labels would be generated for subsequent epochs until training is over, which is equivalent to exploiting them to gradually fine-tune the model.
Moreover, inspired by self-paced learning \cite{kumar2010self}, instead of generating all the pseudo-labels at once time, APL would progressively generate more confident ones of them.
Specifically, negative pseudo-labels are generated with a sample proportion $\theta^\prime\% = \theta\% / (T_\text{t} - T_\text{w})$ in each epoch, where $T_\text{t}$ is the total training epoch. 
For the $c$-th class, APL firstly sorts the predicted probability set $U_c = \{f^{(n)}_c|y^{(n)}_c = 0, n=1,2,...,N\}$ in ascending order. 
Then, APL assigns negative pseudo-labels of the $c$-th class to the images with the $\theta^\prime\%$ lowest predicted probabilities in $U_c$. 
For an partially labeled image $\mathbf{x}^{(n)}$,  negative pseudo-labels are generated by following
\begin{equation}
    {y}^{(n)}_c \triangleq -\mathbbm{1}_{[f^{(n)}_c \in U_c^\prime]}, \quad
    \forall c \in \{c | {y}^{(n)}_c = 0\},
    \label{eq:PL}
\end{equation}
where $U_c^\prime \subset U_c$ denotes the $\theta^\prime\%$ lowest predicted probability subset of $U_c$, and ${y}^{(n)}_c \triangleq -1$ means that $\mathbf{x}^{(n)}$ is assigned the negative pseudo-label of the $c$-th class. Moreover, instead of assigning hard labels to images, APL records $s^{(n)}_c \triangleq f^{(n)}_c$ as soft labels for them to mitigate the effect of noisy negative pseudo-labels. Thus, the loss term of the negative pseudo-label is
\begin{align}
    \mathcal{L}_*(f^{(n)}_c, s^{(n)}_c) = 
          s^{(n)}_c        \log(f^{(n)}_c)
        + (1 - s^{(n)}_c)  \log(1 - f^{(n)}_c)
      .
      \label{eq:PL_loss}
\end{align}
As for the remaining unannotated labels, 
we still provide supervision signals for them by entropy maximization.
Finally, adopting $\beta$ as a hyperparameter to down-weight $\mathcal{L}_*$, EM loss combined with APL is formulated as
\begin{align}
    \mathcal{L}_\text{EM+APL} (\mathbf{f}^{(n)}, \mathbf{y}^{(n)}) = 
    -  \frac{1}{C} \sum_{c=1}^{C}
    [
             & \mathbbm{1}_{[y^{(n)}_c = 1]}  \log(f^{(n)}_c)
    +          \mathbbm{1}_{[y^{(n)}_c = 0]}  \alpha H(f^{(n)}_c)  \notag \\
    +        & \mathbbm{1}_{[y^{(n)}_c = -1]} \beta \mathcal{L}_*(f^{(n)}_c, s^{(n)}_c)
    ].
    \label{eq:EM_PL}
\end{align}


\section{Experiments}
\label{sec:4}

\subsection{Experimental Setup}
\noindent \textbf{Datasets.} 
Since there are no existing datasets totally in line with the setting of single positive multi-label learning (SPML), following \cite{cole2021multi}, we use popular large-scale multi-label datasets to simulate “single positive” datasets by discarding annotations, which can guarantee access to all ground-truth labels for analyzing training phenomena and evaluating performance.
Specifically, after withholding 20\% of the training images for validation, we randomly select one positive label to keep and treat the other labels are unannotated ones for each training image, which is performed once for each dataset. Note that the validation and test sets remain fully labeled.
We use PASCAL VOC 2012 (VOC) \cite{everingham2011pascal}, MS-COCO 2014 (COCO) \cite{lin2014microsoft}, NUS-WIDE (NUS) \cite{chua2009nus}, and CUB-200-2011 (CUB) \cite{wah2011caltech} in our experiments. 
More dataset descriptions are contained in Appendix~\ref{sec:C.1}.

\noindent \textbf{Implementation Details.} 
For fair comparison, we follow the main implementation of \cite{cole2021multi}. Specifically, ResNet-50 \cite{he2016deep} pretrained on the ImageNet \cite{russakovsky2015imagenet} is adopted as the backbone, followed by global average pooling \cite{lin2013network} and a $2048 \times C$ fully connection layer that outputs the predicted probabilities of $C$ classes. All input images are resized to 448 $\times$ 448, and the training images are horizontally flipped with a probability of 0.5 for data augmentation. For each method, we conduct a grid search of batch sizes in $\{8, 16\}$ and learning rates in $\{1e-2, 1e-3, 1e-4, 1e-5\}$, and then select the hyperparameters with the best mean average precision (mAP) on the validation set. Moreover, we use Adam \cite{kingma2014adam} as our optimizer to train all models for 10 epochs, and early stopping is performed when mAP on the validation set descends. Each experiment runs three times, and the mean and standard deviations of mAP are reported. 

\noindent \textbf{Comparing Methods.} 
We compare our method to the following methods: 
AN loss (assuming-negative loss),
EntMin (entropy minimization regularization) \cite{grandvalet2004semi},
Focal loss \cite{lin2017focal},
ASL  (asymmetric loss) \cite{ridnik2021asymmetric},
ROLE (regularized online label estimation) \cite{cole2021multi}, and
ROLE+LI (ROLE combined with the “LinearInit”) \cite{cole2021multi}.
Besides, we also compare our method to the baseline (i.e. AN loss) with the following improvement:
DW (down-weighting $\mathcal{L}_{-}$),
L1R/L2R ($l_1$/$l_2$ regularization),
LS (label smoothing), and
N-LS (label smoothing for only $\mathcal{L}_{-}$).
Moreover, we also report the performance of the models trained with binary cross-entropy (BCE) loss on the fully labeled dataset and on a special partially labeled dataset (i.e. 1 P. \& All N. in Table~\ref{tab:exp_results}). 
To simulate the lacking of positive supervision, BCE loss would ignore all unannotated labels  (i.e. their losses are not computed) in the latter case.
Refer to Appendix~\ref{sec:C.2} and \ref{sec:C.3} for the details of the comparing methods and the hyperparameters of each method respectively.

\begin{table}[t]
  \centering
  \setlength{\tabcolsep}{1pt}
  \renewcommand{\arraystretch}{1.2}
  \caption{Experimental results of our method and the comparing methods on four SPML benchmarks. The best performance of the methods in the SPML setting (i.e. 1 P. \& 0 N.) is marked in bold and the second best is marked in italic}
  \resizebox{\linewidth}{!}{
    \setlength{\tabcolsep}{1mm}{
    \begin{tabular}{c|c|cccc}
    \hline
    \textbf{Ann. Labels} & \textbf{Methods} & \textbf{VOC} & \textbf{COCO} & \textbf{NUS} & \textbf{CUB} \bigstrut\\
    \hline
    All P. \& All N. & BCE loss & 89.42$\pm$0.27 & 76.78$\pm$0.13 & 52.08$\pm$0.20 & 30.90$\pm$0.64 \bigstrut[t]\\
    1 P. \& All N. & BCE loss & 87.60$\pm$0.31 & 71.39$\pm$0.19 & 46.45$\pm$0.27 & 20.65$\pm$1.11 \bigstrut[b]\\
    \hline
    \multirow{11}[2]{*}{1 P. \& 0 N.} & AN loss & 85.89$\pm$0.38 & 64.92$\pm$0.19 & 42.27$\pm$0.56 & 18.31$\pm$0.47 \bigstrut[t]\\
          & DW    & 86.98$\pm$0.36 & 67.59$\pm$0.11 & 45.71$\pm$0.23 & 19.15$\pm$0.56 \\
          & L1R   & 85.97$\pm$0.31 & 64.44$\pm$0.20 & 42.15$\pm$0.46 & 17.59$\pm$1.82 \\
          & L2R   & 85.96$\pm$0.36 & 64.41$\pm$0.24 & 42.72$\pm$0.12 & 17.71$\pm$1.79 \\
          & LS    & 87.90$\pm$0.21 & 67.15$\pm$0.13 & 43.77$\pm$0.29 & 16.26$\pm$0.45 \\
          & N-LS  & 88.12$\pm$0.32 & 67.15$\pm$0.10 & 43.86$\pm$0.54 & 16.82$\pm$0.42 \\
          & EntMin & 53.16$\pm$2.81 & 32.52$\pm$5.55 & 19.38$\pm$3.64 & 13.08$\pm$0.15 \\
          & Focal loss & 87.59$\pm$0.58 & 68.79$\pm$0.14 & 47.00$\pm$0.14 & 19.80$\pm$0.30 \\
          & ASL   & 87.76$\pm$0.51 & 68.78$\pm$0.32 & 46.93$\pm$0.30 & 18.81$\pm$0.48 \\
          & ROLE  & 87.77$\pm$0.22 & 67.04$\pm$0.19 & 41.63$\pm$0.35 & 13.66$\pm$0.24 \\
          & ROLE+LI & 88.26$\pm$0.21 & 69.12$\pm$0.13 & 45.98$\pm$0.26 & 14.86$\pm$0.72 \bigstrut[b]\\
    \hline
    \multirow{2}[2]{*}{1 P. \& 0 N.} & EM loss & \textit{89.09$\pm$0.17} & \textit{70.70$\pm$0.31} & \textit{47.15$\pm$0.11} & \textit{20.85$\pm$0.42} \bigstrut[t]\\
          & EM loss+APL & \textbf{89.19$\pm$0.31} & \textbf{70.87$\pm$0.23} & \textbf{47.59$\pm$0.22} & \textbf{21.84$\pm$0.34} \bigstrut[b]\\
    \hline
    \end{tabular}%
    }}
    
    \label{tab:exp_results}
\end{table}

\subsection{Results and Discussion}
The experimental results on four SPML benchmarks are reported in Table~\ref{tab:exp_results}. 
It can be observed that the assumption of AN loss (i.e. assuming all unannotated labels are negative) causes a significant performance drop, e.g. 9.81\% and 12.59\% mAP decrements than being trained with full annotations (i.e. All P. \& All N.) on NUS and CUB respectively. 
When adopting some improvement on AN loss, the impact of this assumption can be mitigated. Down-weighting $\mathcal{L}_{-}$ of Eq.~\ref{eq:AN} can help to achieve competitive performance on CUB, and label smoothing is more useful on VOC. Note that adopting label smoothing for only assumed negative labels can achieve better performance, which verifies the importance of learning from annotated positive labels. Whereas, a common regularization technique, i.e. $l_1$/$l_2$ regularization, seems to under-perform in the SPML setting. 

As for the other comparing methods, Focal loss and ASL can achieve good performance on NUS and CUB, which can reduce the effect of the dominance of assumed negative labels but still can not address this issue well.
The state-of-the-art method of SPML (i.e. ROLE), which adopts a jointly trained label estimator to perform label estimation during training, has competitive performance in SPML, e.g. achieving 88.77\% mAP on VOC. 
Moreover, the “LinearInit” training fashion can be integrated with ROLE to further improve its performance. However, this state-of-the-art method still achieves a poor classification result on CUB.
It is worth noting that entropy minimization regularization, which is the opposite of our entropy maximization (EM) loss, achieves badly poor and unstable results on all datasets. It demonstrates that this widely adopted method of semi-supervised learning does not applicable to SPML, since only positive labels are annotated for multi-label training images in this special setting.

It can be observed that our EM loss, which adopts entropy maximization for unannotated labels, outperforms the existing methods on all four SPML benchmarks, e.g. achieving 89.09\% and 70.70\% mAP on VOC and COCO respectively.
Note that ROLE uses a learnable label matrix to act as the label estimator, which significantly increases the memory consumption during training. 
Moreover, this method requires access to the average number of positive labels, which is practically unavailable in the SPML setting. 
Compared with ROLE, our EM loss is cost-free and does not introduce any extra learnable parameters in light of its simplicity. 
Besides, the model trained with EM loss can even outperform the model trained with additional supervision (i.e. 1 P. \& All N.) in three benchmarks (i.e. VOC, NUS, and CUB). 
Moreover, it can also be observed that asymmetric pseudo-labeling (APL) can further boost the performance of the model trained with EM loss, and finally our EM loss+APL achieves state-of-the-art results on all four benchmarks.
Especially, EM loss+APL can achieve a 6.98\% mAP increment than ROLE+LI on CUB, and achieves 89.19\% mAP on VOC, which even approaches the result of being trained with full annotations.

\begin{table}[t]
    \caption{
    Left: Experimental results of AN loss, ROLE, and EM loss on VOC with different backbones.
    Right: Precision (averaged in 3 runs) of pseudo-labels generated by APL cooperating with AN and EM loss on four multi-label datasets
    }
    \resizebox{0.55\linewidth}{!}{
    \setlength{\tabcolsep}{1mm}{
    \begin{tabular}{c|ccc}
    \hline
    \textbf{Methods} & \textbf{ResNet-34} & \textbf{ResNet-50} & \textbf{ResNet-101} \bigstrut\\
    \hline
    AN loss & 84.60$\pm$0.29 & 85.89$\pm$0.38 & 86.59$\pm$0.19 \bigstrut[t]\\
    ROLE  & 85.44$\pm$0.16 & 87.77$\pm$0.22 & 88.37$\pm$0.07 \\
    EM loss & \textbf{86.58$\pm$0.25} & \textbf{89.09$\pm$0.17} & \textbf{88.80$\pm$0.11} \bigstrut[b]\\
    \hline
    \end{tabular}%
    }}
    \hfill
    \resizebox{0.44\linewidth}{!}{
    \setlength{\tabcolsep}{1mm}{
    \begin{tabular}{c|cccc}
    \hline
    \textbf{Losses} & \textbf{VOC} & \textbf{COCO} & \textbf{NUS} & \textbf{CUB} \bigstrut\\
    \hline
    AN loss  & 96.44\% & 96.32\% & 98.16\% & 89.82\% \bigstrut\\
    EM loss  & \textbf{99.57}\% & \textbf{99.67}\% & \textbf{99.89}\% & \textbf{91.58}\% \bigstrut\\
    \hline
    \end{tabular}%
    }}
  \label{tab:diff_backbone}
\end{table}

\begin{table}[t]
  \centering
    \caption{Experimental results of APL and the pseudo-labeling variants on CUB. $\checkmark$ indicates that the corresponding technique are adopted. Note that the first row denotes that pseudo-labeling is not performed and the last row is our APL}
    \resizebox{0.65\linewidth}{!}{
    \setlength{\tabcolsep}{1mm}{
    \begin{tabular}{cccc|c}
    \hline
    \textbf{High Prop.} & \textbf{Soft Label} & \textbf{DW} & \textbf{Pos. PL} & \textbf{mAP} \bigstrut\\
    \hline
          &       &       &       & 20.85$\pm$0.42 \bigstrut[t]\\
          & $\checkmark$     & $\checkmark$     &       & 20.95$\pm$0.34 \\
    $\checkmark$     &       & $\checkmark$     &       & 20.99$\pm$0.38 \\
    $\checkmark$     & $\checkmark$     &       &       & 21.00$\pm$0.44 \\
    $\checkmark$     & $\checkmark$     & $\checkmark$     & $\checkmark$     & 20.82$\pm$0.68 \bigstrut[b]\\
    \hline
    $\checkmark$     & $\checkmark$     & $\checkmark$     &       & \textbf{21.84$\pm$0.34} \bigstrut\\
    \hline
    \end{tabular}
    }}
  \label{tab:APL}%
\end{table}

\begin{figure*}[t]
    \centering
    \includegraphics[width=1\linewidth]{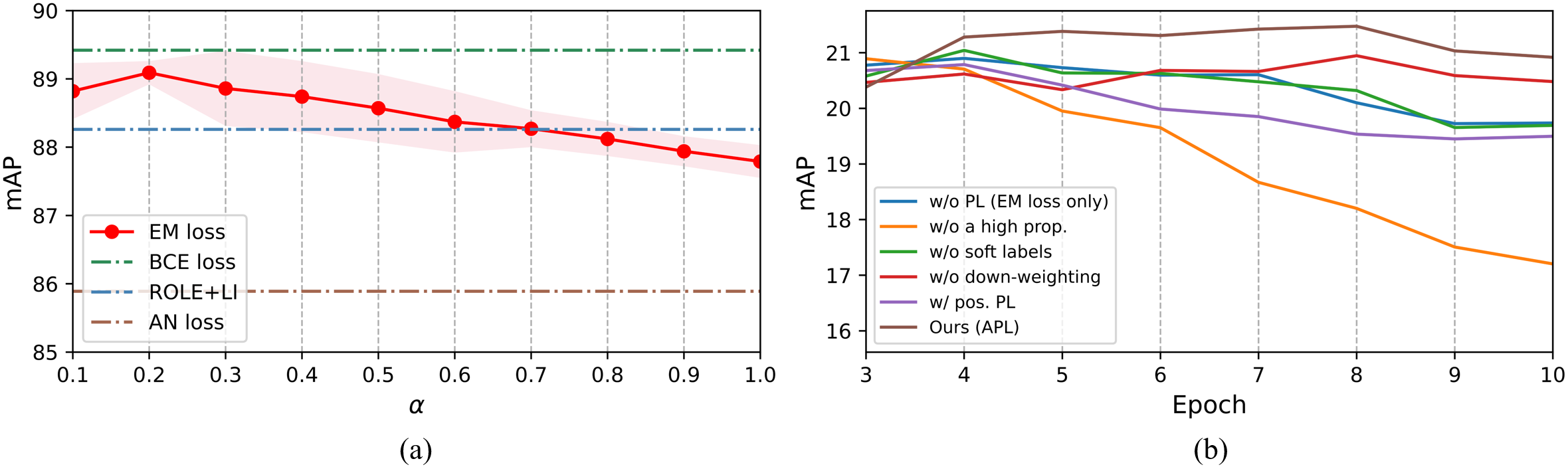}
    \caption{
    (a): Experimental results (averaged in 3 runs) of EM loss with different $\alpha$ on VOC.
    (b): Performance curves (averaged in 3 runs) of APL and its variants on CUB. Note that the x-axis ticks start by 3 since the first 3 epochs are for warm-up
    }
    \label{fig:ABL}
\end{figure*}

\subsection{Ablation Study}
\label{sec:4.3}
\noindent \textbf{Robustness of EM Loss to $\alpha$.} 
To investigate how performance is affected by $\alpha$, we present the experimental results of EM loss with different $\alpha$ in Fig.~\ref{fig:ABL}(a). 
The performance has a peak at $\alpha=0.2$ on VOC and an improper $\alpha$ might cause a performance drop. 
However, the variation range of mAP remains slight and EM loss can still achieve relatively high performance with different $\alpha$. 
Especially, EM loss with all different $\alpha$ can still outperform AN loss, and even approaches the result of being trained with full annotations (i.e. BCE loss) when $\alpha=0.2$, showing that performance can be improved by selecting a proper $\alpha$.

\noindent \textbf{Effectiveness of EM Loss on Different Backbones.}
To study the impact of the special gradient regime of EM loss when the scale of the deep model varies, we consider ResNet with different depths (i.e. ResNet-34, ResNet-50, and ResNet-101) as backbones to be trained with EM loss. As reported in Table.~\ref{tab:diff_backbone}, EM loss achieves the best performance on these backbones as we expect. However, the performance of EM loss on ResNet-50 is even better than on ResNet-101, which can be reserved as an interesting work for further exploration.

\noindent \textbf{Evaluation of APL.}
We perform detailed ablation study for APL (see Fig.~\ref{fig:ABL}(b) and Table~\ref{tab:APL}). Specifically, we remove or add several key components of APL to obtain the following five variants: 
1) w/o pseudo-labeling: the model is trained only with EM loss.
2) w/o a high sample proportion: $\theta\%$ is set to 10\%.
3) w/o soft labels: the pseudo-labeled samples are assigned hard pseudo-labels.
4) w/o down-weighting: $\beta$ is set to 1.
5) w/ positive pseudo-labeling: positive pseudo-labeling is also performed in the same way as negative pseudo-labeling, and its sample proportion is set to 10\%. 
The results on the test set of CUB show that APL can achieve more stable training and each component adopted by APL contributes to performance improvement (see Appendix~\ref{sec:D} for detailed analysis). 
Moreover, we report the precision of pseudo-labels generated by APL (see Table.~\ref{tab:diff_backbone}, we disable early stopping for fair comparison), which shows that APL can generate pseudo-labels with high precision ($\approx 99\%$) in EM loss on most datasets.
Though unannotated labels are already assumed as negative ones in AN loss, we also try adopting APL in AN loss to select potential ground-truth negative labels for reference, showing the effectiveness of asymmetric-tolerance strategies and the capability of EM loss to reduce false negative predictions. 

\subsection{Further Analysis}
\begin{figure*}[t]
    \centering
    \includegraphics[width=1\linewidth]{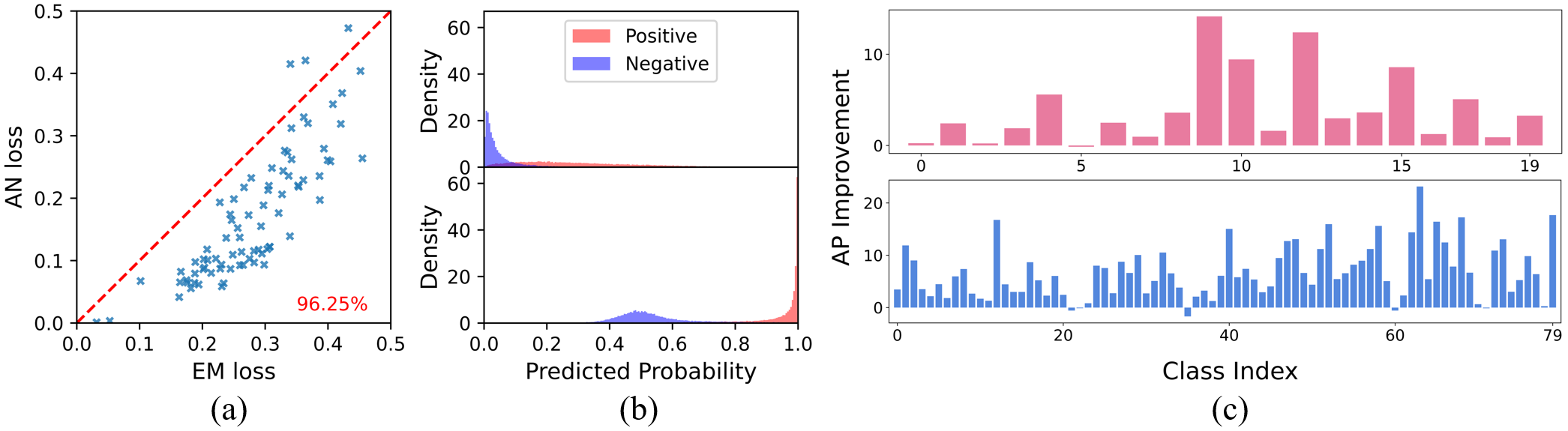}
    \caption{
    (a): Wasserstein distances between the distributions of the predicted probabilities for unannotated positive and negative labels on the training set of COCO, where each point indicates one class. 
    We present the proportion of points that are under the red line. 
    (b): Densities of predicted probabilities on the most common “person” class of the test set of COCO, produced by the models trained with AN (\textit{top}) and EM (\textit{bottom}) loss.
    (c): Class-wise AP improvement of EM loss on VOC (\textit{top}) and COCO (\textit{bottom})
    }
    \label{fig:anal}
\end{figure*}

\noindent \textbf{Class-wise Performance Improvement.} 
To investigate whether EM loss is beneficial to most classes of the “single positive” datasets, we report class-wise average precision (AP) improvement of EM loss (over AN loss) on the test sets of VOC and COCO.
As shown in Fig.~\ref{fig:anal}(c), our EM loss can improve performance on most classes, 
or more specifically, on 95\% classes of VOC and 93.75\% classes of COCO, which demonstrates its general effectiveness.

\noindent \textbf{Distinguishability of Model Predictions.} 
A well-generalizing model should be able to produce informative predictions for unannotated labels, i.e. the predicted probabilities of unannotated positive and negative labels are more distinguishable.
The divergences between these predicted probabilities can be quantitatively calculated by the Wasserstein distance. As shown in Fig.~\ref{fig:anal}(a), the model trained with EM loss usually achieves a much larger Wasserstein distance, which means that it can yield more distinguishable predictions on most classes.
Furthermore, we also visualize the predicted probabilities on the test set. For the most common “person” class of COCO, as shown in Fig.~\ref{fig:anal}(b), EM loss also contributes to more distinguishable predictions for the test images. 
We also provide more visualization results of other classes in Appendix~\ref{sec:F.2}.  

\begin{figure*}[t]
    \centering
    \includegraphics[width=1\linewidth]{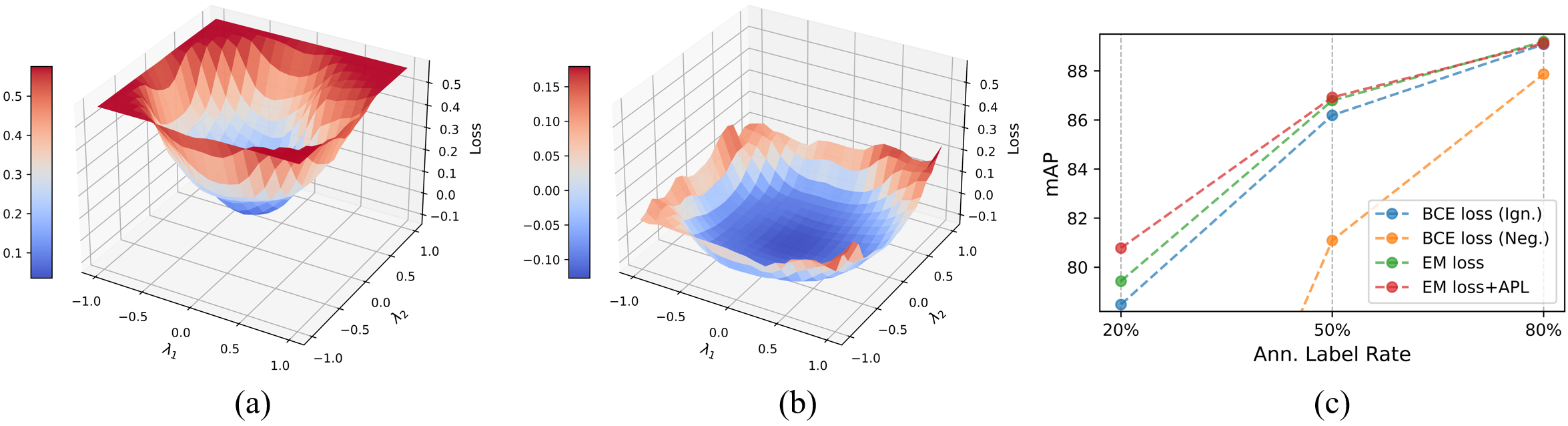}
    \caption{
    (a)\&(b): 3D visualization of loss landscapes around the minima of the models trained with AN and EM loss on VOC respectively.
    Note that we set the z-axis to the same scale for comparing the sharpness.
    (c): Comparison of our method and two baselines in the MLML setting. We randomly discard a percentage of annotations per image as done in \cite{durand2019learning} to simulate MLML datasets
    }
    \label{fig:landscape_and_MLML}
\end{figure*}

\begin{figure*}[t]
    \centering
    \includegraphics[width=1\linewidth]{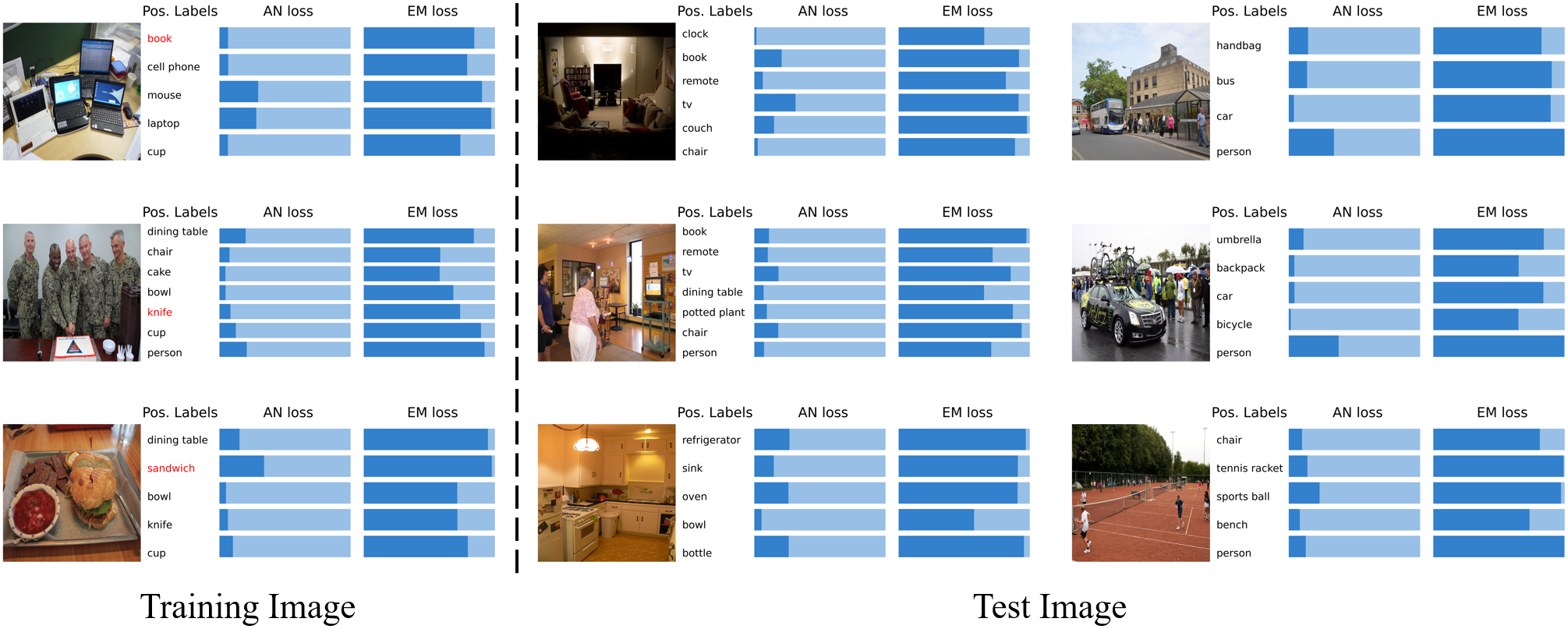}
    \caption{
    Examples of training  and test images of COCO. 
    Note that the annotated positive labels are marked in red for each training image, and the blue bars indicate the predicted probabilities produced by the models trained with AN and EM loss
    }
    \label{fig:viz_image_example}
\end{figure*}

\noindent \textbf{Generalization Evaluation by Loss Landscapes.}
Considering the sharpness of loss minima, we can further explore why EM loss helps to better generalization. 
Given established principles of statistics and information theory, the study in \cite{achille2018emergence} shows that a flat minimum has lower mutual information between training data and model parameters and thus leads to better generalization. Several previous works \cite{hochreiter1997flat,keskar2016large,neyshabur2017exploring} also show that a flat minimum can make the converged model generalize well. Using the technique in \cite{li2017visualizing}, we visualize the loss landscapes in Fig.~\ref{fig:landscape_and_MLML}.
It can be observed that the model trained with AN (resp. EM) loss would converge to a sharp (resp. flatter) minimum, which further explains why the model trained with EM loss can acquire better generalization.

\noindent \textbf{Qualitative results.}
As shown in Fig.~\ref{fig:viz_image_example}, we can observe that the model trained with AN loss would produce low predicted probabilities for almost all unannotated positive labels, especially for small and rare objects in the images. 
Besides, the predictions of the test images also show that AN loss would severely damage model generalization.
However, the model trained with EM loss can produce relatively confident positive predictions for positive labels of training and test images, even for those small and rare objects, which further demonstrates the remarkable effectiveness of EM loss.  

\noindent \textbf{Performance in a More General Scenario.}
Our work focuses on learning from single positive labels, but we are curious about whether it can perform well in a more general setting, i.e. multi-label learning with missing labels (MLML). 
Thus, we report the preliminary experimental results in Fig.~\ref{fig:landscape_and_MLML}. 
Note that the baselines are BCE loss with unannotated labels being ignored or regarded as negative ones and EM loss acts like BCE loss with unannotated labels being supervised by entropy maximization in MLML. 
We find that our method can still achieve performance improvement, which shows that it can be generalized to other similar tasks and is well worth further exploration in the future.

\section{Conclusions}
In this paper, we aim to solve an extreme variant of multi-label learning called single positive multi-label learning (SPML) which is a challenging but under-explored problem. 
Instead of assuming all unannotated labels are negative as traditionally done, we choose to treat them from an alternative perspective, i.e. acknowledging they are unknown. Hence, we propose entropy-maximization (EM) loss and asymmetric pseudo-labeling (APL) for SPML. Experiments show that our method achieves state-of-the-art results on all four benchmarks. 
We believe that our method can serve as a stronger baseline for SPML and expect this paper can motivate more future work of weakly supervised learning.

\subsubsection{Acknowledgements.}
This work is supported by 
the National Key R\&D Program of China (2022YFE0200700), 
the National Natural Science Foundation of China (Project No. 62006219 and No. 62072452), 
the Natural Science Foundation of Guangdong Province (2022A1515011579), 
and the Hong Kong Innovation and Technology Fund (Project No. GHP/110/19SZ and ITS/170/20).

%
%
\bibliographystyle{splncs04}
\bibliography{references}

\clearpage

\title{Acknowledging the Unknown for Multi-label Learning with Single Positive Labels
(Supplementary Material)}

\titlerunning{Acknowledging the Unknown}
\authorrunning{Zhou et al.}

\author{
Donghao Zhou\inst{1,2} \and
Pengfei Chen\inst{3} \and
Qiong Wang\inst{1} \and 
Guangyong Chen\inst{4}\textsuperscript{(\Letter)} \and
Pheng-Ann Heng\inst{1,5}
}

\institute{
Guangdong Provincial Key Laboratory of Computer Vision and Virtual Reality Technology, Shenzhen Institute of Advanced Technology, \\ Chinese Academy of Sciences, Shenzhen, China \\
\email{dh.zhou@siat.ac.cn} \and
University of Chinese Academy of Sciences, Beijing, China \and
Tencent Technology, Shenzhen, China \and
Zhejiang Lab, Hangzhou, China \\
\email{gychen@zhejianglab.com} \and
The Chinese University of Hong Kong, Hong Kong, China
}

\appendix

\maketitle

\setcounter{figure}{6}
\setcounter{table}{3}
\setcounter{equation}{8}

\section{Derivation of the Gradient Equations}
In this section, we provide detailed derivation of the gradients of AN and EM loss (i.e. Eq.~\ref{eq:AN_G} and Eq.~\ref{eq:EM_G}). following the notations in the main paper, the gradients of $\mathcal{L}_{+}$, $\mathcal{L}_{-}$ and $\mathcal{L}_{\varnothing}$ for the logit $g$ are
\begin{align}
      \frac{\partial \mathcal{L}_{+}}{\partial g} 
  & = \frac{\partial \mathcal{L}_{+}}{\partial p} 
      \frac{\partial p}{\partial g} 
    = - \frac{1}{p} \cdot p (1 - p)
    = p - 1  \notag \\
  & = \frac{1}{1 + e^{-g}} - 1
   = \frac{-e^{-g}}{1 + e^{-g}},
\end{align}
\begin{align}
      \frac{\partial \mathcal{L}_{-}}{\partial g} 
  & = \frac{\partial \mathcal{L}_{-}}{\partial p} 
      \frac{\partial p}{\partial g}
    = \frac{1}{1 - p} \cdot p (1 - p)  \notag \\
  & = p
    = \frac{1}{1 + e^{-g}},
\end{align}
\begin{align}
      \frac{\partial \mathcal{L}_{\varnothing}}{\partial g} 
  & = \frac{\partial \mathcal{L}_{\varnothing}}{\partial p} \frac{\partial p}{\partial g} 
    = \alpha [\log{p}-\log{(1-p)}] \cdot p(1-p)  \notag \\
  & = \alpha \log{\frac{p}{1 - p}} \cdot p(1-p) \notag \\
  & = - \alpha \log{e^{-g}} \cdot \frac{1}{1+e^{-g}}\frac{e^{-g}}{1+e^{-g}}  \notag \\
  & = \frac{- \alpha  e^{-g} \log{e^{-g}}}{(1 + e^{-g})^2}.
\end{align}

\section{Additional Empirical Evidence for EM Loss}
\label{sec:B}
In Sec.~\ref{sec:3.2}, we have claimed that the gradient regime of EM loss leads to three behaviours beneficial to model training: 
1) Learning from annotated labels preferentially. 
2) Mitigating the effect of label noise. 
3) Maintaining confident positive predictions.
In this section, we provide more empirical evidence to further verify these three advantages respectively, and thus further demonstrate the effectiveness of EM loss. 
As done in the main paper, for empirical analysis, we consider AN loss as the baseline of SPML, and adopt AN and EM loss in model training respectively. Note that the experimental setup and the hyperparameters of each method are the same as those in benchmark experiments.

\begin{figure*}[t]
    \centering
    \includegraphics[width=1\linewidth]{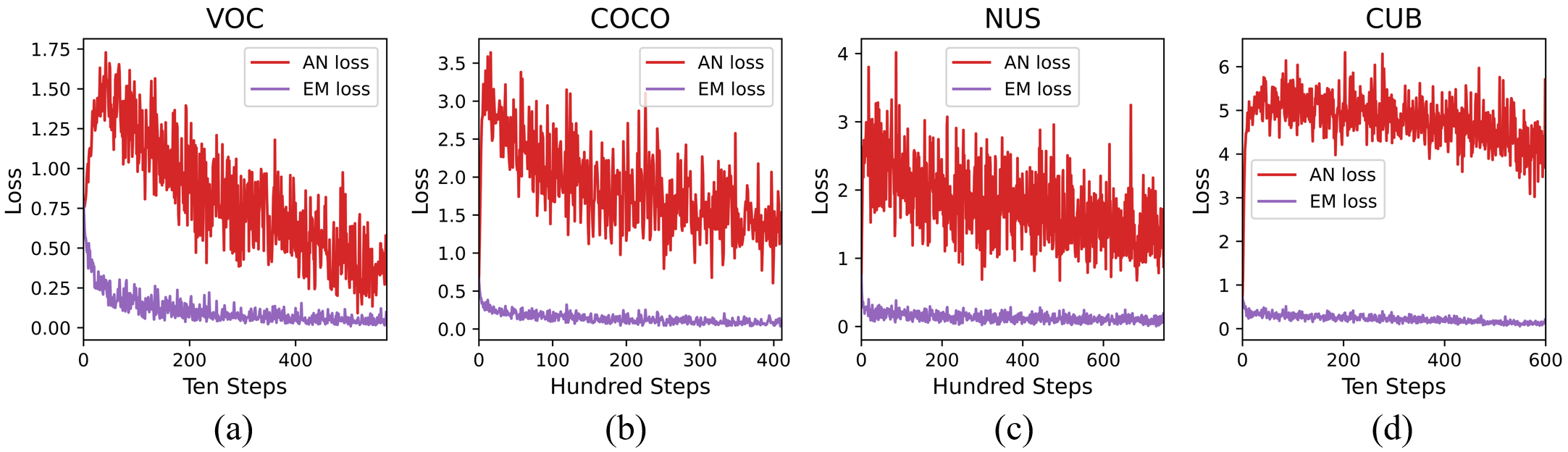}
    \caption{
    Training losses of annotated positive labels (i.e.  $\mathcal{L}_{+}$) on all four datasets from the models trained with AN and EM loss, where $\alpha$ of EM loss is set to the corresponding value shown in Table~\ref{tab:hp}
    }
    \label{fig:loss_pos_all}
\end{figure*}

\begin{figure*}[t]
    \centering
    \includegraphics[width=1\linewidth]{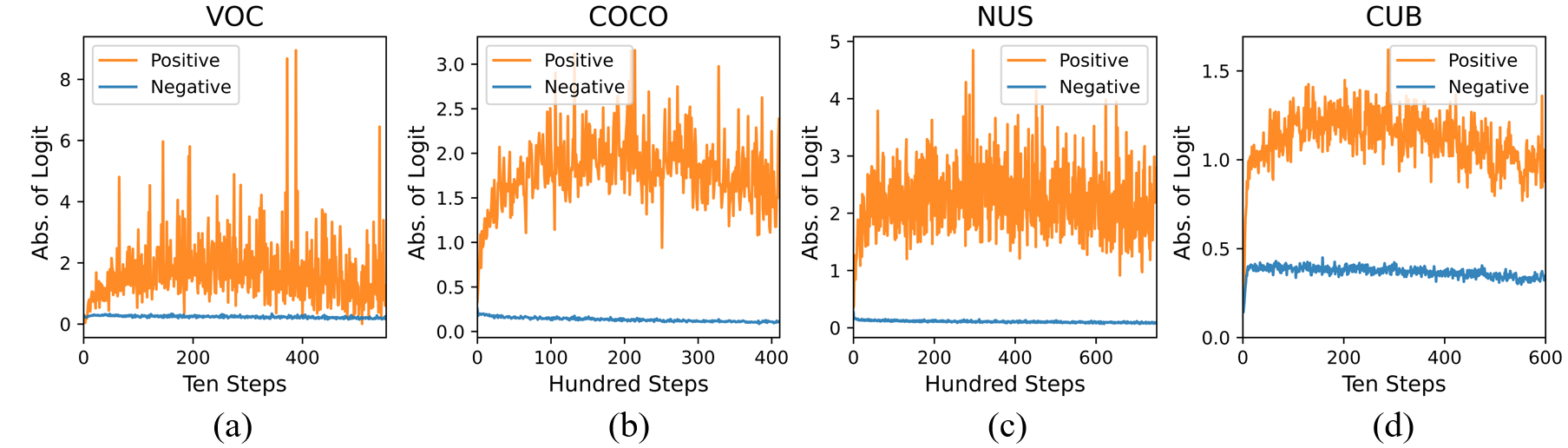}
    \caption{
    Absolute values of the logits for unannotated positive and negative labels, produced by the model trained with EM loss on all four datasets
    }
    \label{fig:logits_all}
\end{figure*}

\subsection{Learning from Annotated Labels Preferentially}
We have claimed that the model training would be dominated by assumed negative labels when adopting AN loss, whereas EM loss can lead the model to preferentially learn from annotated positive labels.
To verify this, we present the training losses of annotated positive labels  (i.e.  $\mathcal{L}_{+}$) on VOC in the main paper, where $\alpha$ of EM loss is set to 1.
We are curious about if this improvement of EM loss can perform well on all four datasets when $\alpha$ is set to a more proper value.
Thus, we provide more visualization of $\mathcal{L}_{+}$ for comparison, where $\alpha$ is set to the corresponding value shown in Table~\ref{tab:hp}.
As shown in Fig.~\ref{fig:loss_pos_all}, on all four datasets, $\mathcal{L}_{+}$ of AN loss would increase in early training, since the model trained with AN loss preferentially focuses on fitting the numerous assumed negative labels.
However, $\mathcal{L}_{+}$ of EM loss can gradually decrease and is more stable, 
since the gradients of EM loss for unannotated labels would be relatively low in early training (see Fig.~\ref{fig:method}(a)).
Moreover, we have also claimed that EM loss tends to keep the predictions of unannotated labels ambiguous.
To empirically observe this, we visualize the absolute values of the logits produced by the model trained with EM loss. 
In Fig.~\ref{fig:logits_all}, it can be observed that EM loss would keep near-zero logits for numerous unannotated negative labels, which results in small gradients for them throughout training.
As for the relatively large results of unannotated positive labels, we would discuss them in Sec.~\ref{sec:C.3}.

\begin{figure*}[t]
    \centering
    \includegraphics[width=0.97\linewidth]{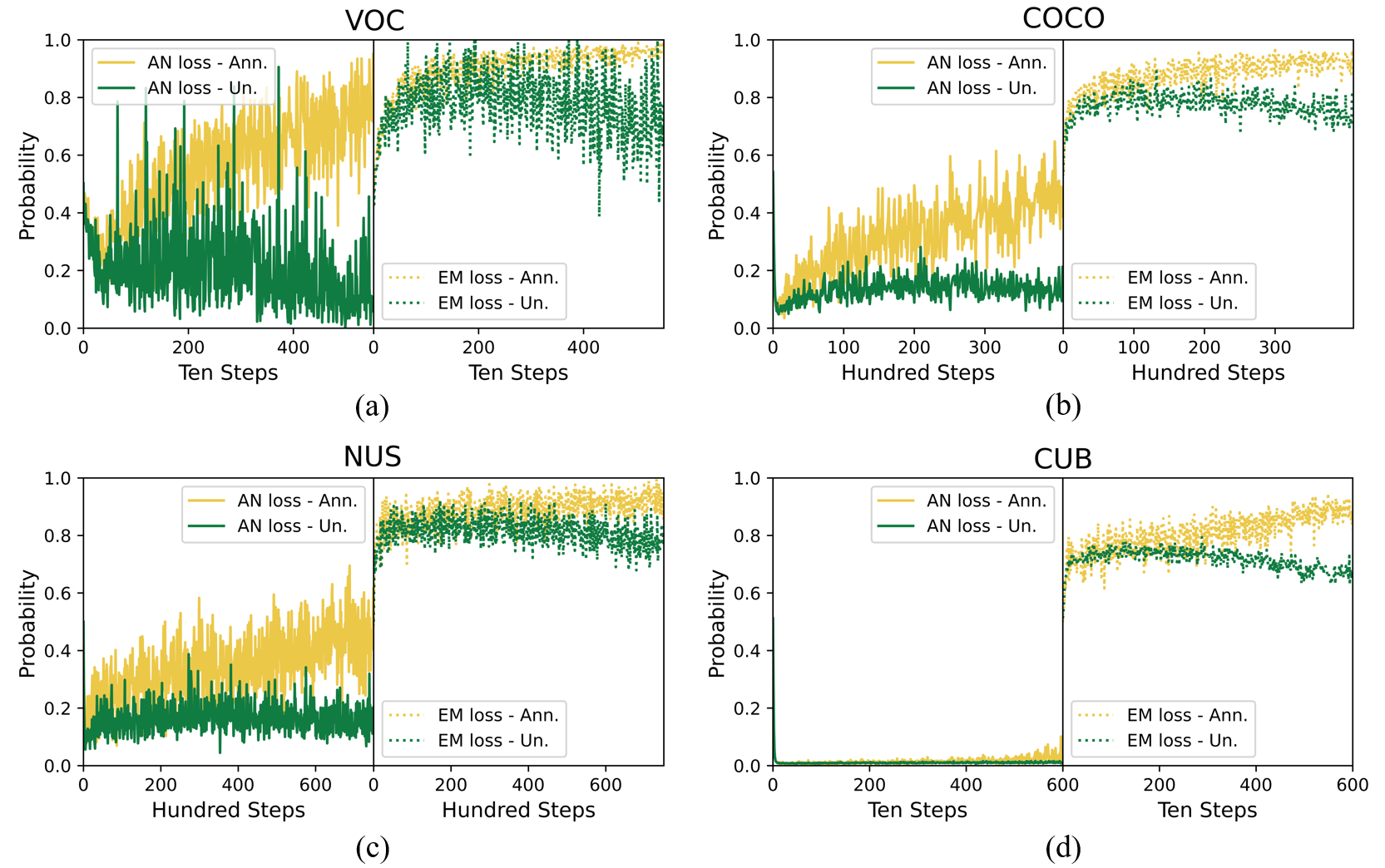}
    \caption{
    Predicted probabilities for annotated and unannotated positive labels, produced by the models trained with AN (\textit{left}) and EM (\textit{right}) loss on all four datasets
    }
    \label{fig:prob_all}
\end{figure*}

\subsection{Mitigating the Effect of Label Noise}
In Sec.~3.2, we have also claimed that the model trained with AN loss would suffer from the false negative labels (i.e. unannotated positive labels which are assumed as negative ones), whereas EM loss can mitigate the effect of label noise. 
Trained with AN loss, the model would be confused by annotated and unannotated positive labels, which results in unconfident and even incorrect predictions for positive labels. 
To demonstrate this, we present the predicted probabilities for annotated and unannotated positive labels, which are produced by the models trained with AN and EM loss.
As shown in Fig.~\ref{fig:prob_all}, the model trained with AN loss would produce low predicted probabilities for positive labels, especially for the unannotated ones.
However, the model trained with EM loss can produce relatively confident positive predictions for both unannotated and annotated positive labels, since EM loss does not introduce any false negative labels and is capable of encouraging the model to learn from annotated positive ones.

\subsection{Maintaining Confident Positive Predictions}
Different from AN loss, EM loss can maintain confident positive predictions for unannotated labels due to its special gradient regime. This claim can be verified by visualizing the logits for unannotated positive labels, since confident positive predictions are always associated with large logits for unannotated positive labels. 
For instance, a logit of 1 (resp. 2, 3) corresponds to a predicted probability of 0.73  (resp. 0.88, 0.95).
As shown in Fig.~\ref{fig:logits_all}, compared to the logits of annotated negative labels, the model trained with EM loss would produce relatively large logits for unannotated positive labels, since EM loss would not over-suppress confident positive predictions by providing large gradients for them. This difference between the logits of unannotated positive and negative labels shows that EM loss can indeed maintain confident positive predictions, instead of keeping near-zero logits for all unannotated labels without distinction.

\section{Details of the Experimental Setup}
In this section, we provide more details of the experimental setup, including dataset descriptions and hyperparameter tuning and selection, to ensure the fairness and reproducibility of our experiments.

\subsection{Datasets}
\label{sec:C.1}
The following large-scale multi-label datasets are used in our experiments: PASCAL VOC 2012 (VOC) \cite{everingham2011pascal}, MS-COCO 2014 (COCO) \cite{lin2014microsoft}, NUS-WIDE (NUS) \cite{chua2009nus}, and CUB-200-2011 (CUB) \cite{wah2011caltech}. 
VOC consists of 5,717 training images and 20 classes, and we report test results on its official validation set with 5,823 images. 
COCO contains 82,081 training images and 80 classes, and we also report test results on its official validation set with 40,137 images. 
NUS consists of 81 classes and contains 150,000 training images and 60,260 testing images collected from Flickr.
Instead of re-crawling the NUS images as done in \cite{cole2021multi}, we use the \textit{ official version} of NUS in our experiments, which has less manual intervention and thus is fairer.
CUB is divided into 5,994 training images and 5,794 test images, consisting of 312 classes (i.e. binary attributes of birds).
For reference purposes, we summarize the statistics of the datasets in Table~\ref{tab:dataset}, which shows the diversity of these four popular multi-label datasets.

\begin{table}[t]
  \centering
    \caption{Statistics of the datasets, including the number of classes, the number of images on the split datasets, and the number of ground-truth positive and negative labels per image on the training sets}
    \setlength{\tabcolsep}{2mm}{
    \begin{tabular}{c|c|cccc} 
    \hline
    \multicolumn{2}{c|}{\textbf{Statistics}} & \textbf{VOC} & \textbf{COCO} & \textbf{NUS} & \textbf{CUB} \bigstrut\\
    \hline
    \multicolumn{2}{c|}{\# Classes} & 20    & 80    & 81    & 312 \bigstrut\\
    \hline
    \multirow{3}[2]{*}{\# Images} & Training  & 4574  & 65665 & 120000 & 4795 \bigstrut[t]\\
          & Validation & 1143  & 16416 & 30000 & 1199 \\
          & Test  & 5823  & 40137 & 60260 & 5794 \bigstrut[b]\\
    \hline
    \multirow{2}[2]{*}{\# Labels Per Training Image} & Positive & 1.46  & 2.94  & 1.89  & 31.4 \bigstrut[t]\\
          & Negative & 18.54 & 77.06 & 79.11 & 280.6 \bigstrut[b]\\
    \hline
    \end{tabular}%
    }
    \label{tab:dataset}
\end{table}

\subsection{Comparing Methods}
\label{sec:C.2}
We compare our method with the following methods: 
1) AN loss (Eq.~\ref{eq:AN}): The widely recognized baseline of SPML, which assumes all unannotated labels are negative.
2) EntMin \cite{grandvalet2004semi}: A widely adopted method of semi-supervised learning, i.e. entropy minimization regularization, which aims to minimize the entropy of predicted probabilities for unannotated labels.
3) Focal loss \cite{lin2017focal}: An efficient method to handle label imbalance.
4) ASL \cite{ridnik2021asymmetric}: One of the state-of-the-art methods of multi-label classification, which can mitigate the effect of mislabeled samples. 
5) ROLE \cite{cole2021multi}: the state-of-the-art method of SPML, which adopts a label estimator and exploits the average number of positive labels to perform regularization.
6) ROLE+LI \cite{cole2021multi}: ROLE is combined with the “LinearInit” training fashion, i.e. firstly training the model with the backbone being frozen before end-to-end training.
Note that unannotated labels are also assumed as negative ones in Focal loss and ASL.
Besides, we also compare our method to the baseline of SPML with the following improvement:
1) DW: Down-weighting $\mathcal{L}_{-}$ of Eq.~\ref{eq:AN}.
2) L1R/L2R: adopting $l_1$/$l_2$ regularization.
3) LS: Label smoothing for all labels.
4) N-LS: Label smoothing for only assumed negative labels. 

\subsection{Hyperparameters}
\label{sec:C.3}
For each method, method-specific hyperparameters are tuned on all four datasets respectively, and the hyperparameters with the best mAP on validation sets are selected for the final evaluation. The detailed hyperparameter tuning and selection of our experiments are as follows:
\begin{enumerate}[1.]
    \item \textbf{DW}: 
    A hyperparameter tuned in $\{0.01, 0.02, 0.1, 0.2, 0.4, 0.9\}$ is used to down-weight $\mathcal{L}_{-}$ of Eq.~\ref{eq:AN}. 
    Finally, 0.1 is selected for VOC, COCO, and NUS, and 0.02 is selected for CUB.
    \item \textbf{L1R/L2R}: 
    A hyperparameter tuned in $\{1e-9, 1e-8, 1e-7, 1e-6, 1e-5\}$ is used to control the strength of $l_1$/$l_2$ regularization. 
    In L1R, we select $1e-6$ (resp. $1e-7$, $1e-7$, $1e-9$) for VOC (resp. COCO, NUS, CUB).
    In L2R, we select $1e-6$ (resp. $1e-7$, $1e-6$, $1e-8$) for VOC (resp. COCO, NUS, CUB).
    \item \textbf{LS/N-LS}: 
    Label smoothing coefficient is tuned in $\{0.1, 0.2, 0.3\}$. 
    In LS, we select 0.1 for all four datasets.
    In N-LS, we select 0.3 for VOC, and select 0.1 for the other datasets.
    \item \textbf{EntMin} \cite{grandvalet2004semi}: 
    A hyperparameter tuned in $\{0.01, 0.02, 0.1, 0.2, 0.4, 0.9\}$ is used to control the strength of entropy minimization regularization. 
    Finally, we select $0.01$ (resp. $0.9$, $0.4$, $0.4$) for VOC (resp. COCO, NUS, CUB). 
    \item \textbf{Focal loss} \cite{lin2017focal}: 
    There are a focusing parameter $\gamma$ and a balance parameter $\alpha$ in Focal loss. 
    As recommended in \cite{ridnik2021asymmetric}, we set $\gamma=2$, and tune $\alpha \in \{0.25, 0.5, 0.75\}$.
    Finally, we select $\alpha=0.75$ for all four datasets.
    \item \textbf{ASL} \cite{ridnik2021asymmetric}: 
    There are two hyperparameters (i.e. $\gamma_+$ and $\gamma_-$) used to control the focusing levels of positive and negative labels respectively, and a hyperparameter (i.e. $m$) used to act as the proposed probability margin in ASL.
    As done in \cite{ridnik2021asymmetric}, we set $\gamma_+=0$, and tune $\gamma_- \in \{1, 2\}$ and $m \in \{0, 0.05, 0.2\}$ with a grid search. 
    Finally, we select $\gamma_-=2$, $m=0.2$ for VOC, COCO, and NUS, and select $\gamma_-=1$, $m=0$ for CUB.
    \item \textbf{ROLE/ROLE+LI} \cite{cole2021multi}: 
    The experimental results are reproduced by reimplementing the methods exactly following the hyperparameters in \cite{cole2021multi}. 
    \item \textbf{EM loss/APL}: 
    For our method, we tune $\alpha, \beta \in \{0.01, 0.02, 0.1, 0.2, 0.4, 0.9\}$. 
    Moreover, we set $\theta\%=90\%$ for all datasets and empirically select $T_\text{w}$ for each dataset. 
    For convenience, the final hyperparameters of our method are shown in Table~\ref{tab:hp}, including the selected batch sizes and learning rates.
\end{enumerate}

\begin{table}[t]
  \centering
    \caption{Hyperparameters of our method on all four datasets in our experiments}
    \setlength{\tabcolsep}{2mm}{
    \begin{tabular}{c|cccc}
    \hline
    \textbf{hyperparameters} & \textbf{VOC} & \textbf{COCO} & \textbf{NUS} & \textbf{CUB} \bigstrut\\
    \hline
    Batch Size & 8   & 16  & 16  & 8 \bigstrut[t]\\
    Learning Rate & $1e-5$  & $1e-5$  & $1e-5$  & $1e-4$ \\
    $\alpha$ & 0.2   & 0.1   & 0.1   & 0.01 \\
    $\beta$ & 0.02  & 0.9   & 0.2   & 0.4 \\
    $\theta\%$ & 90\%   & 90\%    & 90\%    & 90\%  \\
    $T_\text{w}$ & 5     & 5     & 4     & 3 \bigstrut[b]\\
    \hline
    \end{tabular}%
    }
    \label{tab:hp}
\end{table}

\section{Detailed Analysis for APL}
\label{sec:D}
As an extension to the ablation study of APL in Sec.~4.3, we provide detailed analysis for it in this section to further demonstrate the contribution of the components adopted in APL.
We focus on answering the following key questions:

\begin{enumerate}[]
    \item 
    {\noindent \bf Question 1}: 
    Is a high sample proportion necessary for generating negative pseudo-labels? \\
    {\noindent \bf Answer 1}: 
    In Table~\ref{tab:APL}, it can be observed that generating negative pseudo-labeling with a low sample proportion just leads to a tiny mAP increment. 
    As shown in Fig.~\ref{fig:ABL}(b), generating negative pseudo-labeling with a low sample proportion gradually reduces performance as pseudo-labeling goes on, since the model may be overfitting to few negative pseudo-labels. 
    Thus, adopting a high-tolerance strategy is necessary for generating negative pseudo-labels.
    
    \item 
    {\noindent \bf Question 2}: 
    Is assigning hard labels also able to boost performance? \\
    {\noindent \bf Answer 2}:  
    As shown in Table~\ref{tab:APL}, when assigning hard labels instead of soft ones, pseudo-labeling would not significantly boost performance, since label noise may be contained in the generated negative pseudo-labels. As a solution, soft labels can mitigate this damaging impact and make negative pseudo-labels participate in model training in a more appropriate way, which is beneficial to better performance (see Table~\ref{tab:APL}). 
    
    \item
    {\noindent \bf Question 3}: 
    Does down-weighting contribute to performance improvement? \\
    {\noindent \bf Answer 3}: 
    As shown in Fig.~\ref{fig:ABL}(b), pseudo-labeling without down-weighting (i.e. $\beta$ of Eq.~\ref{eq:EM_PL} is set to 1) can still achieve stable training, whereas properly performing down-weighting for the loss of pseudo-labels can lead to further performance improvement (see Table~\ref{tab:APL}).

    \item
    {\noindent \bf Question 4}: 
    What is the effect of performing positive pseudo-labeling? \\
    {\noindent \bf Answer 4}: 
    As shown in Table~\ref{tab:APL}, it is worth noting that performing similar positive pseudo-labeling would cause a performance drop. Since positive labels are the tiny minority of multi-label annotations, positive pseudo-labeling would introduce a large amount of label noise, even with a small sample proportion. To empirically observe this, we present the precision of positive pseudo-labels in Table~\ref{tab:acc_p}, which shows that performing positive pseudo-labeling can only generate positive pseudo-labels with very low precision due to the positive-negative label imbalance of unannotated labels. Therefore, considering this imbalance, we choose to adopt an extreme low-tolerance strategy for positive pseudo-labels, aiming to avoid introducing any noisy positive pseudo-labels for more stable training (see Fig.~\ref{fig:ABL}(b)).
\end{enumerate}

\begin{table}[t]
  \centering
    \caption{Precision (averaged in 3 runs) of positive pseudo-labels generated by the similar positive pseudo-labeling on four multi-label datasets. Note that the sample proportion for positive pseudo-labeling is set to 10\% }
    \setlength{\tabcolsep}{4mm}{
    \begin{tabular}{c|cccc}
    \hline
          & \textbf{VOC} & \textbf{COCO} & \textbf{NUS} & \textbf{CUB} \bigstrut\\
    \hline
    \textbf{Precision} & 15.66\% & 15.29\% & 9.18\% & 19.34\% \bigstrut\\
    \hline
    \end{tabular}%
    }
  \label{tab:acc_p}%
\end{table}

\section{Evaluation with Other Metrics}
In multi-label learning, mAP is the primary metric to evaluate model performance. 
To further verify the effectiveness of our method, we perform an additional evaluation with two metrics (i.e. micro-F1 and macro-F1). 
Specifically, we compare EM loss with AN loss in the SPML setting, and report the performance of BCE loss on the fully labeled datasets. 
As commonly done, the thresholds for micro-F1 and macro-F1 are set to 0.5 for BCE and AN loss.
Since EM loss tends to keep ambiguous predictions for unannotated labels, the predicted probabilities for negative labels produced by the model trained with EM loss would be near 0.5 (see Fig.~\ref{fig:more_sample_hist1}).
Therefore, for fair comparison, we set the thresholds to 0.75 for EM loss.
As shown in Table~\ref{tab:f1}, AN loss achieves poor performance on all four datasets, especially on CUB (both micro-F1 and macro-F1 are 0).
However, our EM loss can still perform well in the evaluation with these two metrics. For instance, EM loss can achieve 85.36\% micro-F1 on VOC, which even exceeds the results of being trained with full annotations.

\begin{table}[t]
    \centering
    \setlength{\tabcolsep}{1pt}
    \renewcommand{\arraystretch}{1.2} 
    \caption{Experimental results of BCE loss, AN loss, and EM loss on four SPML benchmarks with two additional metrics (i.e. micro-F1 and macro-F1).
    Note that the model trained with BCE loss adopts full annotations for training
    and the best performance of the methods in the SPML setting is marked in bold
    }
    \resizebox{1\linewidth}{!}{
    \setlength{\tabcolsep}{1mm}{
    \begin{tabular}{c|cccccccc}
    \hline
    \textbf{Datasets} & \multicolumn{2}{c}{\textbf{VOC}} & \multicolumn{2}{c}{\textbf{COCO}} & \multicolumn{2}{c}{\textbf{NUS}} & \multicolumn{2}{c}{\textbf{CUB}} \bigstrut\\
    \hline
    \textbf{Methods} & \textbf{Micro-F1} & \textbf{Macro-F1} & \textbf{Micro-F1} & \textbf{Macro-F1} & \textbf{Micro-F1} & \textbf{Macro-F1} & \textbf{Micro-F1} & \textbf{Macro-F1} \bigstrut\\
    \hline
    BCE loss & 85.18 & 82.84 & 76.01 & 71.84 & 68.97 & 50.59 & 47.54 & 22.15 \bigstrut[t]\\
    \hline
    AN loss & 73.24 & 71.66 & 38.03 & 41.85 & 28.24 & 21.36 & 0     & 0 \\
    EM loss & \textbf{85.36} & \textbf{82.78} & \textbf{71.58} & \textbf{66.74} & \textbf{66.83} & \textbf{45.47} & \textbf{43.85} & \textbf{20.38} \bigstrut[b]\\
    \hline
    \end{tabular}
    }}
  \label{tab:f1}
\end{table}

\begin{figure*}[t]
    \centering
    \includegraphics[width=1\linewidth]{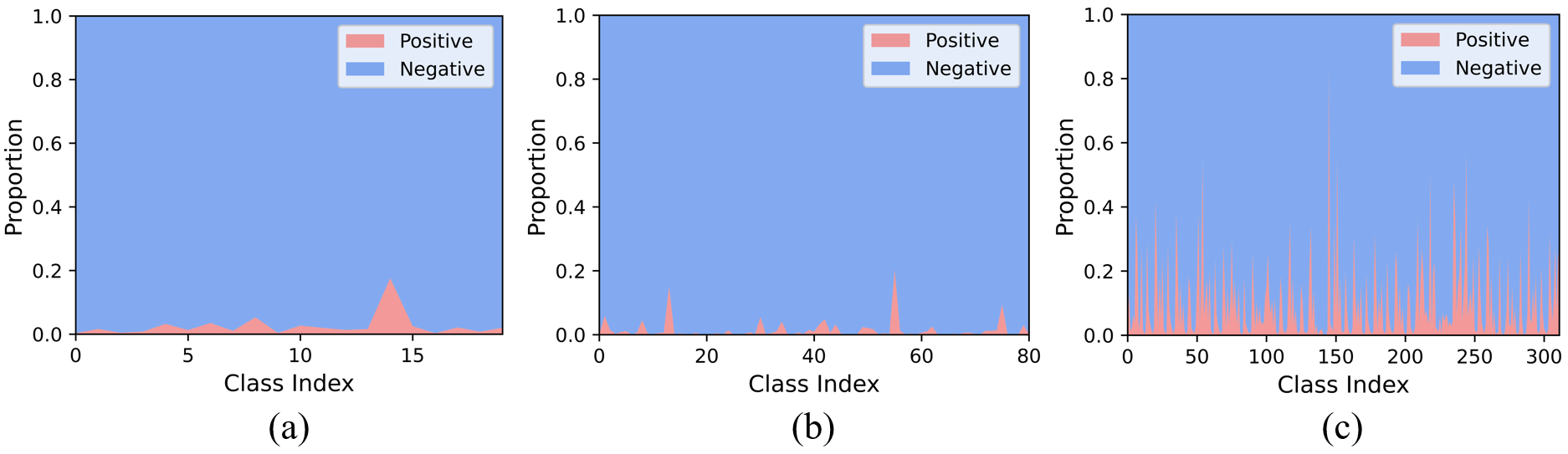}
    \caption{
    Proportions of unannotated positive and negative labels of each class on the other datasets
    }
    \label{fig:imbalance_all}
\end{figure*}

\section{More Illustrative Examples}
In this section, we provide more illustrative examples to support our observations in the main paper, 
including the positive-negative label imbalance of unannotated labels and the distinguishability of model predictions. 

\subsection{Positive-Negative Label Imbalance of Unannotated Labels}
\label{sec:F.1}
As an extension to Fig.~\ref{fig:method}(c), we present the proportions of unannotated positive and negative labels of each class on the other datasets in Fig.~\ref{fig:imbalance_all}, which shows that the positive-negative labels imbalance of unannotated labels is an inherent property of SPML.
As shown in Fig.~\ref{fig:imbalance_all}, it is worth noting that the proportions of unannotated negative labels on some classes are lower than the predefined sample proportion for negative pseudo-labeling (i.e. 90\%).
Fortunately, with a self-paced procedure, it is not often that APL would generate negative pseudo-labels with a sample proportion of 90\% before early stopping, which does not damage the high precision of pseudo-labels generated by APL (see Table~\ref{tab:diff_backbone}).  

\subsection{Distinguishability of Model Predictions}
\label{sec:F.2}
\begin{figure*}[t]
    \centering
    \includegraphics[width=1\linewidth]{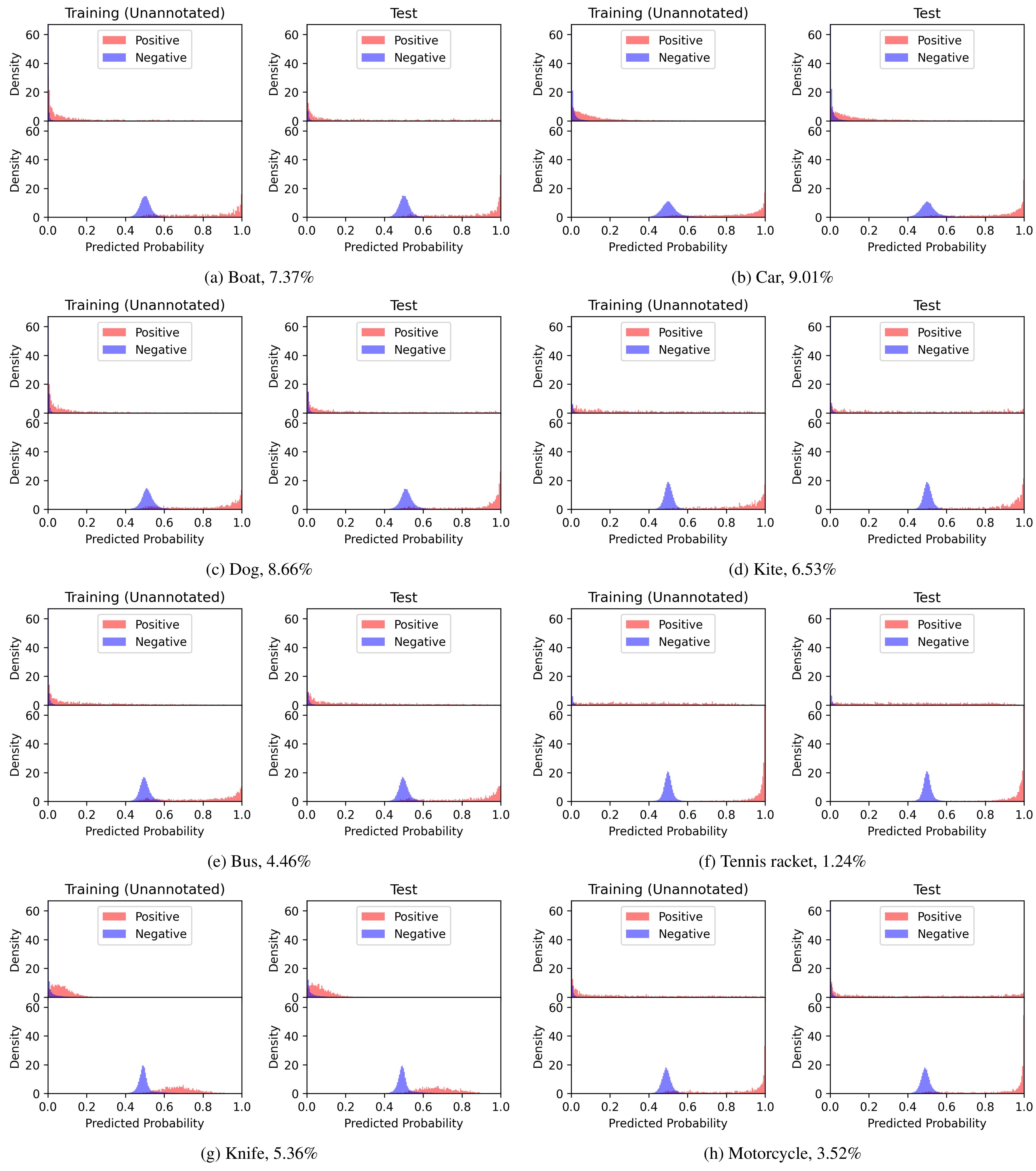}
    \caption{
    Densities of the predicted probabilities for training and test images on more classes of COCO, produced by the models trained with AN (\textit{top}) and EM (\textit{bottom}) loss. 
    Note that we only visualize the \textit{unannotated} labels of training images. 
    The caption of each subfigure contains the class name and the percentage of mAP increment. For clear comparison, we limit the y-axis to the same scale as Fig.~\ref{fig:anal}(b)
    }
    \label{fig:more_sample_hist1}
\end{figure*}

In Fig.~\ref{fig:more_sample_hist1}, we visualize the predicted probabilities for positive and negative labels on more classes of COCO, aiming to further compare the effect of AN and EM loss on the distinguishability of model predictions. 
It can be observed that the model trained with EM loss can produce more distinguishable predictions for positive and negative labels. 
Moreover, we also present the class name and the percentage of mAP increment in the caption of each subfigure in Fig.~\ref{fig:more_sample_hist1}, which shows that distinguishability improvement indeed contributes to model performance as we expect. Especially, as shown in Fig.~\ref{fig:more_sample_hist1}(g), the model trained with EM loss can produce more distinguishable predictions on the “knife” class, even though they are rare and small objects in the images of COCO.

\end{document}